\documentclass{article}

\usepackage{arxiv}
\usepackage{amsmath}
\usepackage{amssymb}
\usepackage{url}
\usepackage{bm}
\usepackage[noend]{algpseudocode}
\usepackage{algorithm,algorithmicx}
\usepackage{varwidth}
\usepackage{float}
\usepackage{graphicx}
\usepackage{siunitx}
\usepackage{layouts}
\usepackage[acronym]{glossaries}
\usepackage{pgfplots}
\usepackage[hidelinks]{hyperref}
\usepackage[caption=false,font=normalsize,labelfont=sf,textfont=sf]{subfig}
\usepackage{placeins}
\usepackage{eso-pic}

\newacronym{dof}{DoF}{degrees of freedom}
\newacronym{sbmp}{SBMP}{sampling-based motion planning}
\newacronym{hri}{HRI}{human-robot interaction}
\newacronym{bohb}{BOHB}{Bayesian Optimization and Hyperband}
\newacronym{ompl}{OMPL}{Open Motion Planning Library}
\newacronym{dbscan}{DBSCAN}{Density Based Spatial Clustering of Applications with Noise}

\title{Parameter Optimization for Manipulator Motion Planning using a Novel Benchmark Set}

\date{}

\author{Carl Gaebert, Sascha Kaden, Benjamin Fischer and Ulrike Thomas
\thanks{All authors are with the Robotics and Human-Machine Interaction Lab, Chemnitz University of Technology, Germany {\tt\footnotesize \{carl.gaebert, sascha.kaden,}
{\tt\footnotesize ulrike.thomas\}@etit.tu-chemnitz.de}%
}
}

\renewcommand{\headeright}{}
\renewcommand{\undertitle}{}
\renewcommand{\shorttitle}{Parameter Optimization for Manipulator Motion Planning using a Novel Benchmark Set}

\hypersetup{
pdftitle={Parameter Optimization for Manipulator Motion Planning using a Novel Benchmark Set},
pdfauthor={DCarl Gaebert, Sascha Kaden, Benjamin Fischer and Ulrike Thomas},
}

\newcommand\copyrighttext{%
  \footnotesize \textcopyright 2023 IEEE. Personal use of this material is permitted. Permission from IEEE must be
  obtained for all other uses, in any current or future media, including
  reprinting/republishing this material for advertising or promotional purposes, creating new
  collective works, for resale or redistribution to servers or lists, or reuse of any copyrighted
  component of this work in other works.}
\newcommand\copyrightnotice{%
\begin{tikzpicture}[remember picture,overlay]
\node[anchor=south,yshift=10pt] at (current page.south) {\fbox{\parbox{\dimexpr\textwidth-\fboxsep-\fboxrule\relax}{\copyrighttext}}};
\end{tikzpicture}%
}

\begin{document}
\maketitle
\copyrightnotice

\begin{abstract}
	Sampling-based motion planning algorithms have been continuously developed for more than two decades.
	Apart from mobile robots, they are also widely used in manipulator motion planning. Hence, these methods play a key role in collaborative and shared workspaces. Despite numerous improvements, their performance can highly vary depending on the chosen parameter setting. The optimal parameters depend on numerous factors such as the start state, the goal state and the complexity of the environment. Practitioners usually choose these values using their experience and tedious trial and error experiments. To address this problem, recent works combine hyperparameter optimization methods with motion planning. They show that tuning the planner's parameters can lead to shorter planning times and lower costs. It is not clear, however, how well such approaches generalize to a diverse set of planning problems that include narrow passages as well as barely cluttered environments.
	In this work, we analyze optimized planner settings for a large set of diverse planning problems. We then provide insights into the connection between the characteristics of the planning problem and the optimal parameters. As a result, we provide a list of recommended parameters for various use-cases. Our experiments are based on a novel motion planning benchmark for manipulators which we provide at \href{https://mytuc.org/rybj}{https://mytuc.org/rybj}.
\end{abstract}


\section{Introduction}
For more than two decades, robot motion planning has been an active field of research.
One prominent group of approaches are \gls{sbmp} algorithms. They are well-suited to address the problem of high-dimensionality which is a main challenge in many robotics applications.
Instead of searching the complete configuration space, these algorithms use random sampling to find a feasible solution. Approaches such as RRT* \cite{Karaman2011rrtstar} and RRT*-Connect \cite{Klemm2015} sample the robot's configuration space to grow a collision-free tree connecting start and goal. While doing so, they also optimize the existing tree to minimize a given cost function.
In contrast to static scenarios, \gls{hri} calls for optimized solutions within a very short planning time.
To this end, many approaches have been presented which increase the performance of sampling-based planners. They address the problem of narrow passages \cite{Hsu2003bridgetest,Amato98}, use adaptive sampling strategies \cite{Rodriguez2008} or even learning-based methods \cite{Ichter2018,Qureshi2019,Cheng2020}.
Despite all these improvements, the performance of the planner is still heavily influenced by its most basic parameter settings, namely the step size $s$ and the goal bias $b_{goal}$.
Moreover, the settings recorded in the original works are often related to mobile robot applications. In industry, practitioners are thus often faced with fine-tuning basic parameter settings by hand for manipulators. A recent line of research therefore aims at automatizing this process through optimization.
In recent years, several approaches have addressed the problem of automatic hyperparameter tuning in \gls{sbmp} for manipulators. In \cite{BurgerAutomated}, the authors utilize a random forest model to tune parameters of several algorithms provided in the \gls{ompl} \cite{sucan2012ompl}. The authors achieved reduced planning times and costs in a few pre-defined pick-and-place scenarios.
Cano et al. investigated several optimization methods while considering the BKPIECE \cite{bkpiece} and RRT-Connect \cite{rrtconnect} algorithms implemented in \gls{ompl}. The authors show that planning time can be reduced by a factor of 4.5. In addition, randomized planning problems are used to show that the tuned parameters generalize well across various setups from the same problem distribution.
\begin{figure}[t!]
    \centerline{
    \subfloat[Trivial environment]{\includegraphics[width=0.48\columnwidth]{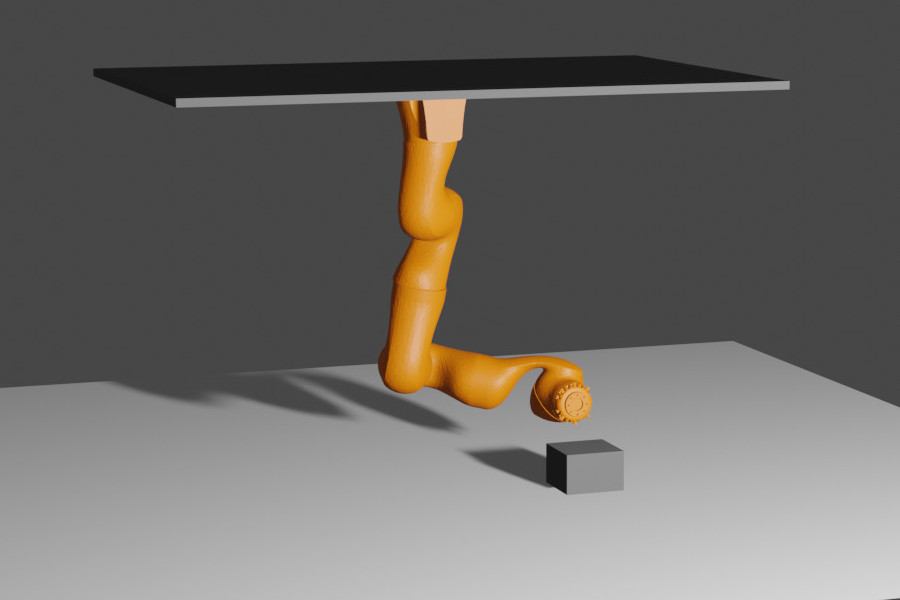} \label{fig:introductiona}}
    \hfil
    \subfloat[Complex environment]{\includegraphics[width=0.48\columnwidth]{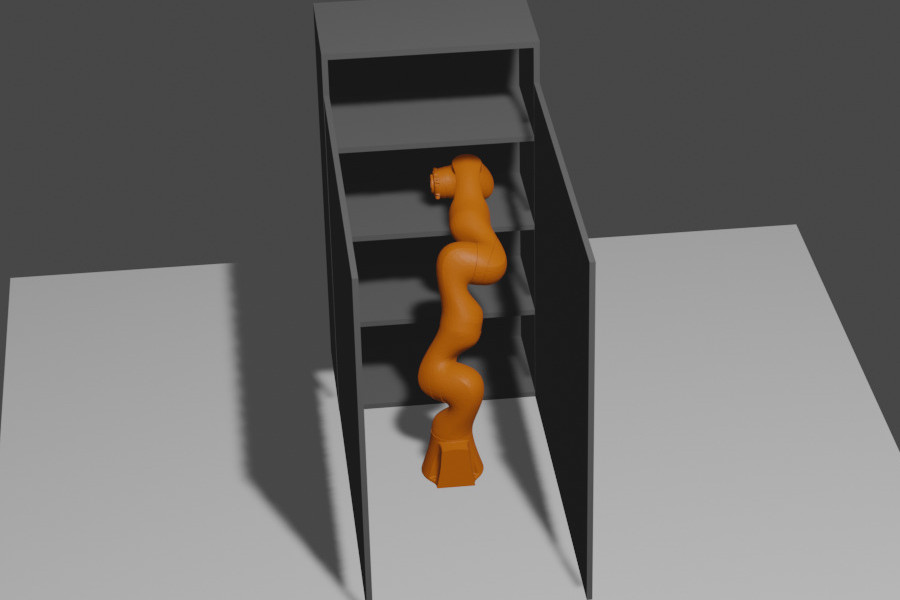} \label{fig:introductionb}}
    }
    \caption{A seven \gls{dof} manipulator operating in two different environments. (a): Almost no obstacles are present which allows for a higher step size and goal bias during sampling. (b): Complex environment with many narrow passages. A smaller goal bias and step size is thus recommended.}

    \label{fig:introduction}
\end{figure}
Moll et al. use \gls{bohb} \cite{BOBHFalkner} to find optimal settings in an extensive search space that includes several planning algorithms and their parameters. Hence, the authors manage to find the best performing approach together with its tuned parameters. For evaluating their approach, the authors use several classes of planning problems such as reaching into a shelf or into a box. The considered optimization criteria are planning and execution speed.
In the same work, the authors also demonstrate that optimizing on a small set of problem classes can lead to good performance in all of them. However, it was not shown to which number of classes and to which variety in complexity this holds. For example, a planner optimized on a trivial problem such as in Fig. \ref{fig:introductiona} may still work well when introducing a few more obstacles to the scene. However, it is likely to fail in very complex environments like the one shown in Fig. \ref{fig:introductionb}.
Furthermore, none of the mentioned works utilize a parallelized planner. It is thus not clear to which extent parameter optimization can improve the results in such a setup. To this end, our work provides the following contributions:


\begin{itemize}
    \item We provide a benchmark dataset for manipulator motion planning consisting of 20 environments and over 200 planning problems of different complexity.
    \item We utilize this dataset to investigate how well optimized parameters generalize across a very diverse set of planning problems.
    \item We analyze the distribution of optimized parameters and connect them to the characteristics of the planning problems. In consequence, we can provide a list of recommendations to practitioners for setting up their planner in a specific scenario.
\end{itemize}

In the remainder of the paper, we first provide a brief summary of the optimization algorithm. Since we use results from stochastic planners, we also review a method for extracting clusters from noisy datasets. Next, we introduce the optimization function used throughout this work. In a subsequent step, we present details of the motion planning benchmark for manipulators in \ref{sec:benchmarks}. Next, we point out the challenges of finding suitable parameters for diverse planning problems. Finally, we experimentally derive a list of recommendations for optimal planning parameters depending on the problem.

\section{Background}
\subsection{Bayesian Optimization and Hyperband}
In \gls{sbmp}, costs and planning time heavily depend on the planner's parameters. A very low goal bias, for example, can lead to a bad performance even in simple environments. Such sensitivity towards hyperparameter choices is also commonly observed in areas of Deep Learning. Therefore, automatic hyperparameter tuning is an active field of research beyond robotics.
Formally, the performance of a motion planning algorithm can be defined as a function $f: \mathcal{X} \rightarrow \mathbb{R}$ of its parameters $x$.
The goal of the optimization procedure is to find the parameters $x^*$ for which $f$ is minimized.
Typically, the function $f$ is expensive to validate. However, it can be approximated by sampling from the parameter space $\mathcal{X}$.
A recent work by Falkner et al. \cite{BOBHFalkner} in this direction combines the strengths of Bayesian Optimization with a Hyperband scheduler \cite{li2017hyperband}. In line with the work of Moll et al. \cite{MollHyperplan}, we utilize this approach to tune the parameters of our motion planning algorithm.
In \gls{bohb} the Hyperband scheduler is used to identify the best out of $n_{trials}$ randomly initialized planner configurations. It does so by using Successive Halving \cite{jamieson16} (as cited in \cite{BOBHFalkner}) and assigning resources to the most promising trials. Instead of relying on random samples, \gls{bohb} uses kernel density estimators to generate new promising trials based on previous results. The resulting increase in efficiency is crucial for our experiments since they require solving hundreds of challenging motion planning problems.

\subsection{Clustering Large and Noisy Datasets}
\gls{sbmp} algorithms rely on stochastic processes and thus deliver non-deterministic solutions. This has to be taken into account while analyzing results from such sources.
When clustering such datasets, one thus has to cope with outliers that do not necessarily belong to a cluster. One established algorithm for such use-cases is \gls{dbscan} \cite{ester199dbscan}. 
The advantage of using this algorithm over K-Means, for example, is that clusters do not have to be of convex shape. Instead, they are viewed as regions with high density which are separated by areas of low density. The method is based on the concept of core samples. A sample is considered a core sample when a number of \textit{min\_samples} are within a distance of $\varepsilon$ to it. Besides core samples, a cluster can also contain samples which are within $\varepsilon$ to one of the core samples. Therefore, the algorithms does not require a number of clusters to look for but a distance metric and the number of core samples. It thus allows for marking too distant samples as outliers.


\section{Methodology}
In \gls{hri} one is typically interested in optimizing for two objectives simultaneously: planning time and costs. The latter can vary depending on the application and may involve state as well as distant costs.
In contrast to works like \cite{MollHyperplan}, we rely on the optimizing planner RRT*-Connect \cite{Klemm2015} only.
For this, we use a custom and highly-parallelized implementation.
This allows us to combine these two objectives into a single objective function which would not be possible using non-optimizing planners such as RRT.

For a single planning problem $p$ from a start configuration $\mathbf{\theta}_{start}$ to a goal configuration $\mathbf{\theta}_{goal}$, we define the combined costs $c(p)$ as follows:
\begin{align}
    c(p) = w_t*t + \frac{c_C}{||\mathbf{\theta}_{start}- \mathbf{\theta}_{goal}||_2}
    \label{eq:costs_single}
\end{align}

The cost consists of the planning time $t$ in seconds and the path length in configuration space $c_C$. The latter is given in radian and normalized using the shortest path length possible. In addition, we weight the influence of the planning time throughout this paper with $w_t = 3$. 

Since we are interested in an approximation of our planner's performance in various workspaces, we consider a whole set $\mathcal{P}$ of $n$ problems. The final objective function $f$ is then defined as
\begin{align}
    f(\mathcal{P}) = \mathcal{Q}_{0.7}\{s_1,...,s_m  \} \label{eq:costs_final}\\
    \text{with } s_j = \sum_{n}c(p_n). \nonumber  
\end{align}
This can be interpreted as follows: We solve all planning problems in $\mathcal{P}$ and sum up the costs calculated using \eqref{eq:costs_single}. The process is repeated $m$ times which results in a distribution of estimates costs for this trial. In line with \cite{MollHyperplan}, we then take the 0.7-quanttile $\mathcal{Q}_{0.7}$ of this distribution as our estimate. This is done to account for the highly stochastic nature and the high variance of solutions obtained by \gls{sbmp} algorithms.
Due to the complexity of the environments, certain parameter configurations might perform very poorly. It is thus necessary to restrict the maximum planning time to $t_{max}$. In cases where no solution was found within $t_{max}$, we use a cost value of $c(p) = 3*t_{max} + 100$.
Moreover, the possible ranges for the goal bias $b_{goal}$ and step size $s$ are significantly different. Using the Euclidean distance for clustering the optimal settings would thus favor clusters with a wide range of goal bias parameters. For this reason, we introduce a distance metric that scales the distance in the goal bias dimension using the ratio of the parameter ranges. The distance between parameter settings $a$ and $b$ are then calculated using the step sizes $s_{a}$ and $s_{b}$ as well as the goal bias settings $b_{goal,a}$ and $b_{goal,b}$ as shown below.
\begin{align}
    d(a,b) = ||s_{a} - s_{b}||_2  + 2.9\times||b_{goal,a} - b_{goal,b}||_2.
    \label{eq:metric}
\end{align}

\section{Motion Planning Benchmark for Manipulators}
\label{sec:benchmarks}
\begin{figure*}[ht]
    \centerline{
    \subfloat[$\mathcal{P}_2$]{\includegraphics[width=0.195\textwidth]{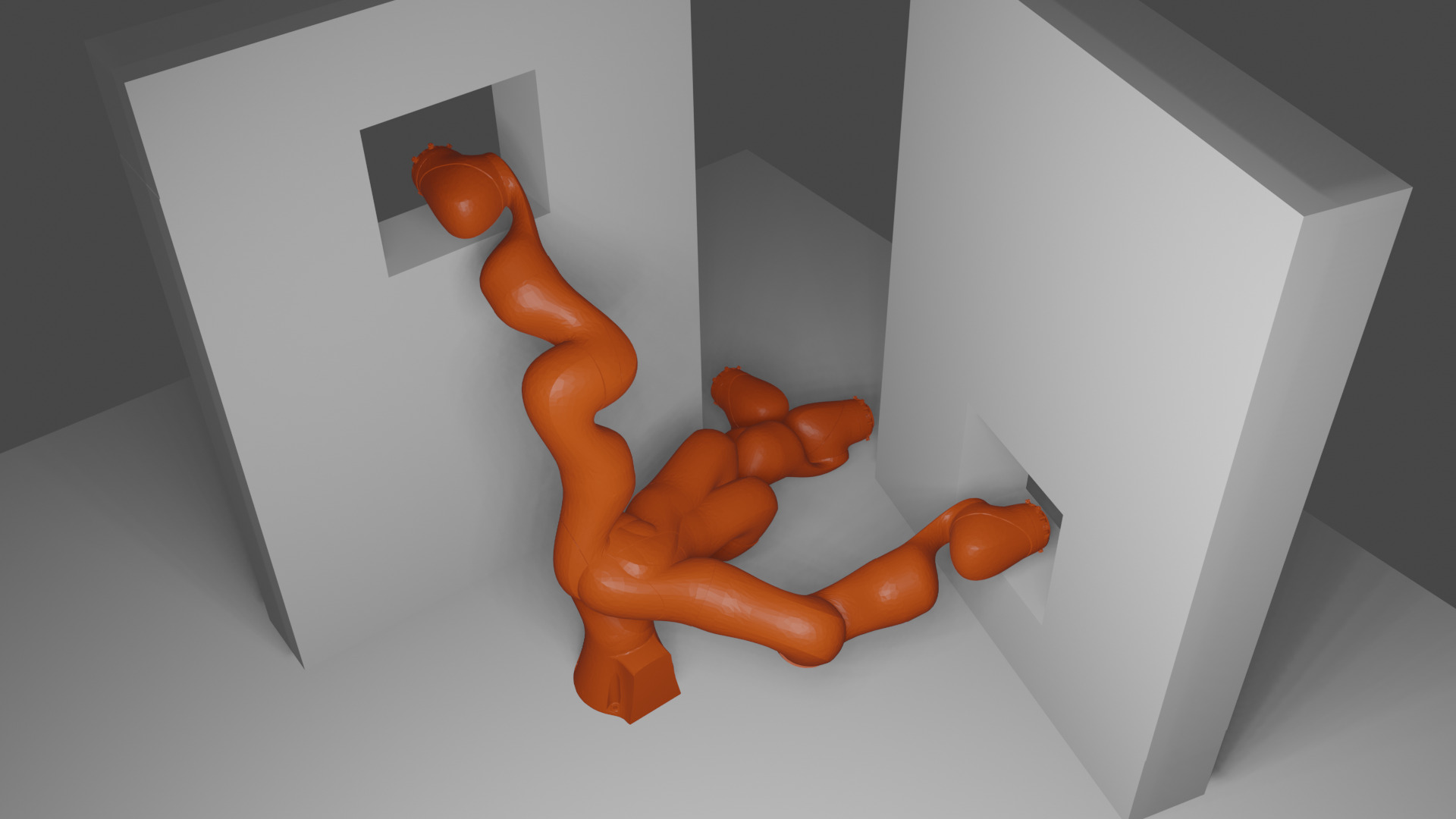}}
    \hfil
    \subfloat[$\mathcal{P}_3$]{\includegraphics[width=0.195\textwidth]{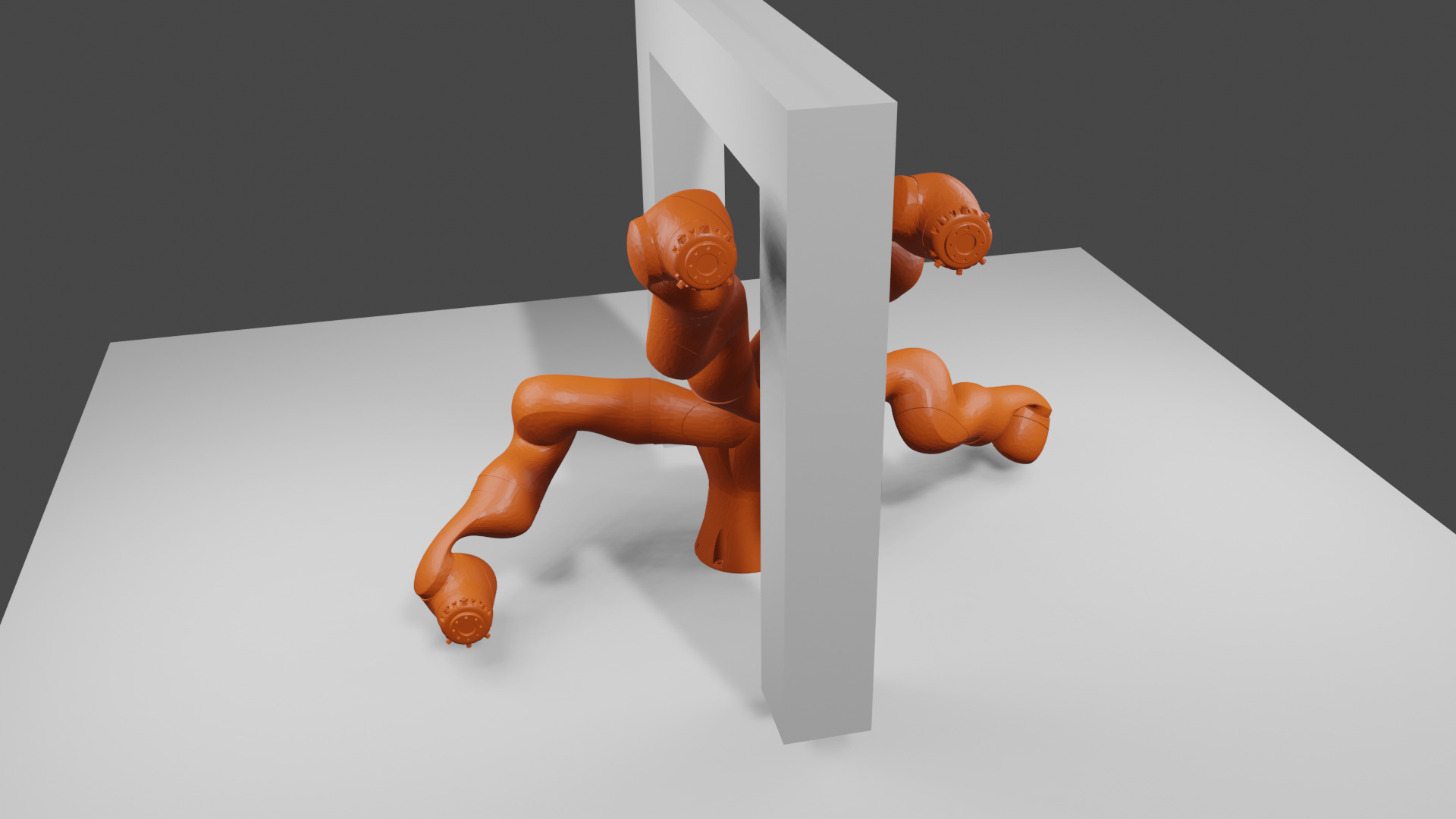}}
    \hfil
    \subfloat[$\mathcal{P}_5$]{\includegraphics[width=0.195\textwidth]{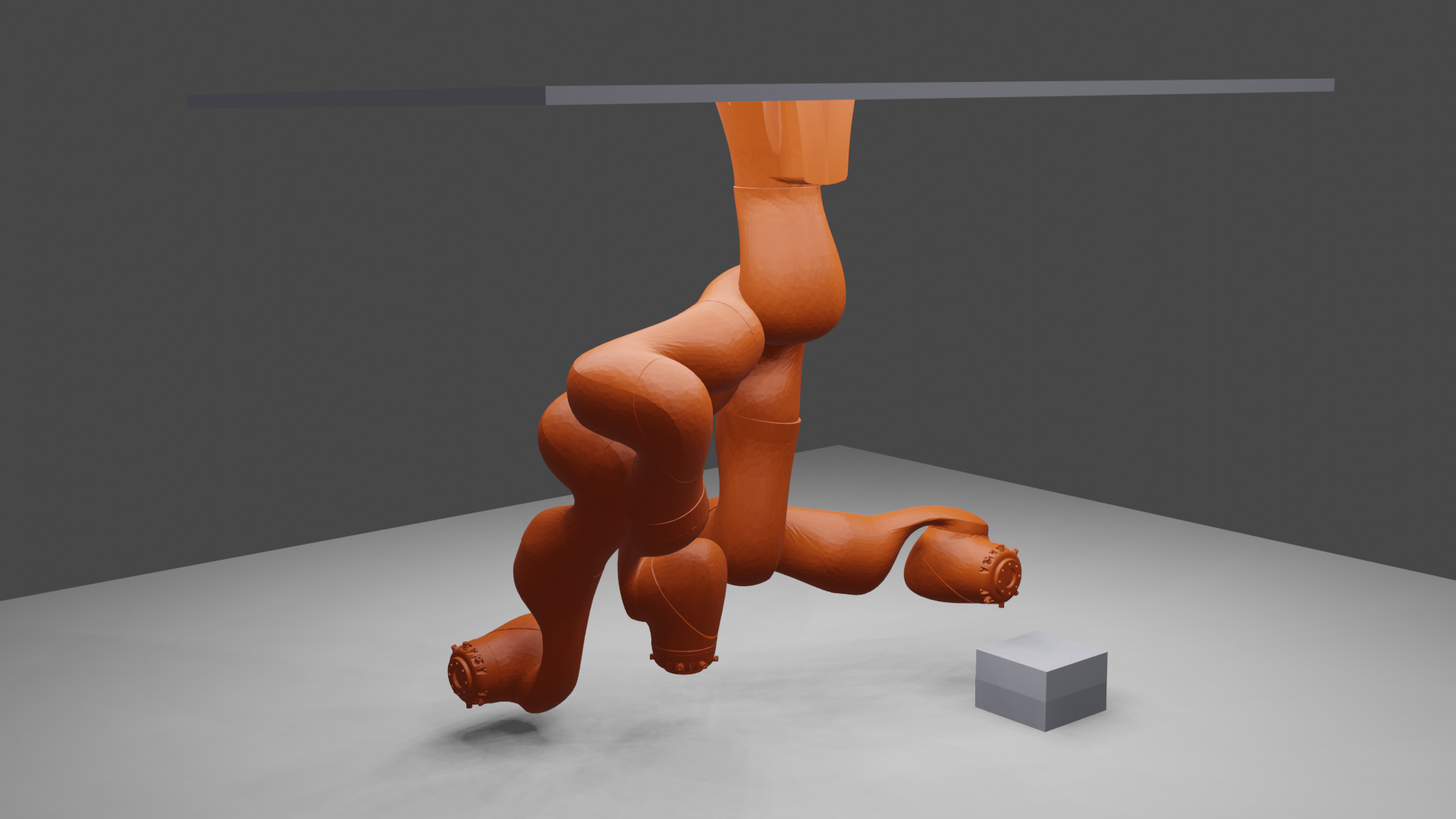}}
    \hfil
    \subfloat[$\mathcal{P}_7$]{\includegraphics[width=0.195\textwidth]{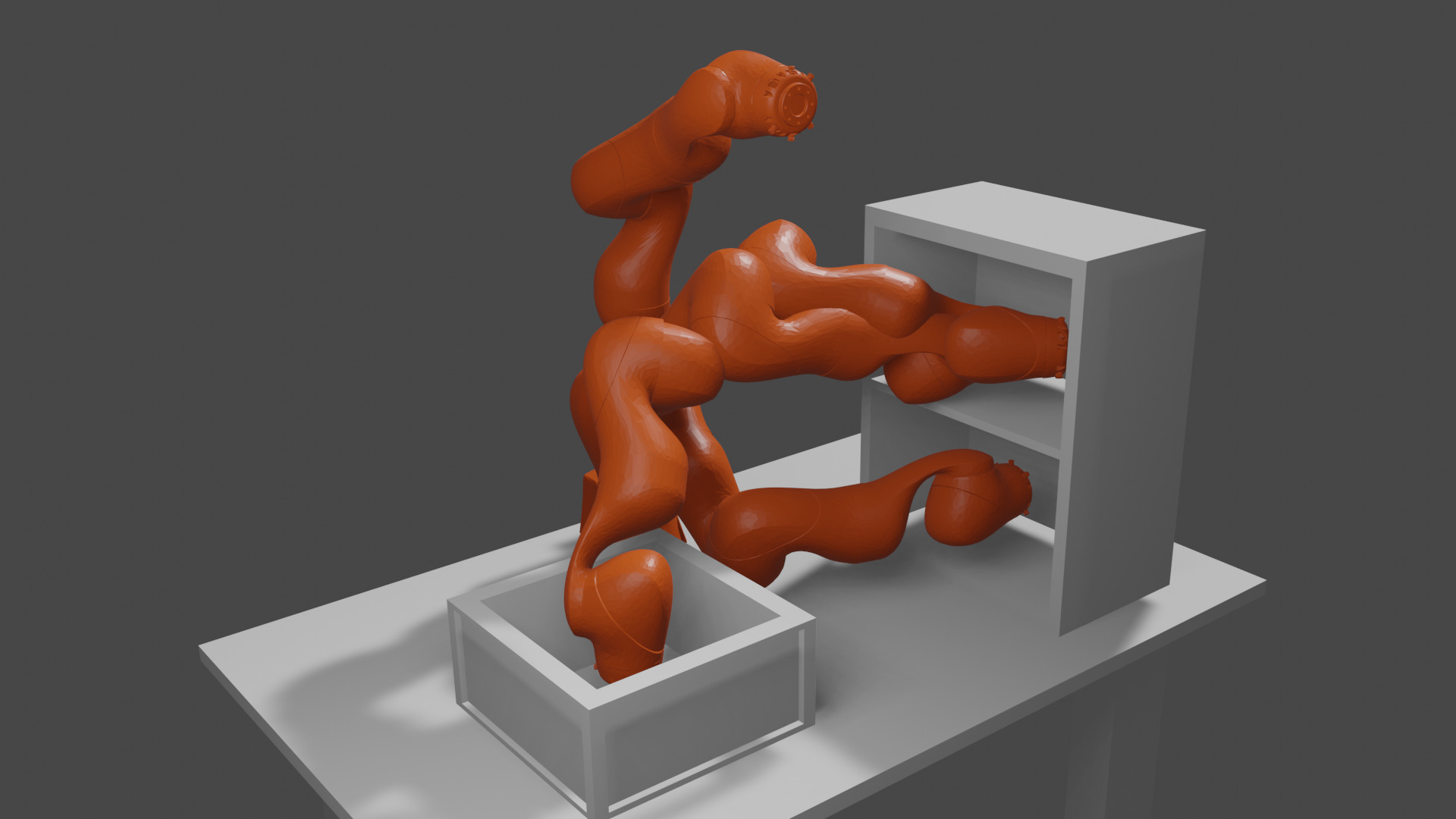}}
    \hfil
    \subfloat[$\mathcal{P}_8$]{\includegraphics[width=0.195\textwidth]{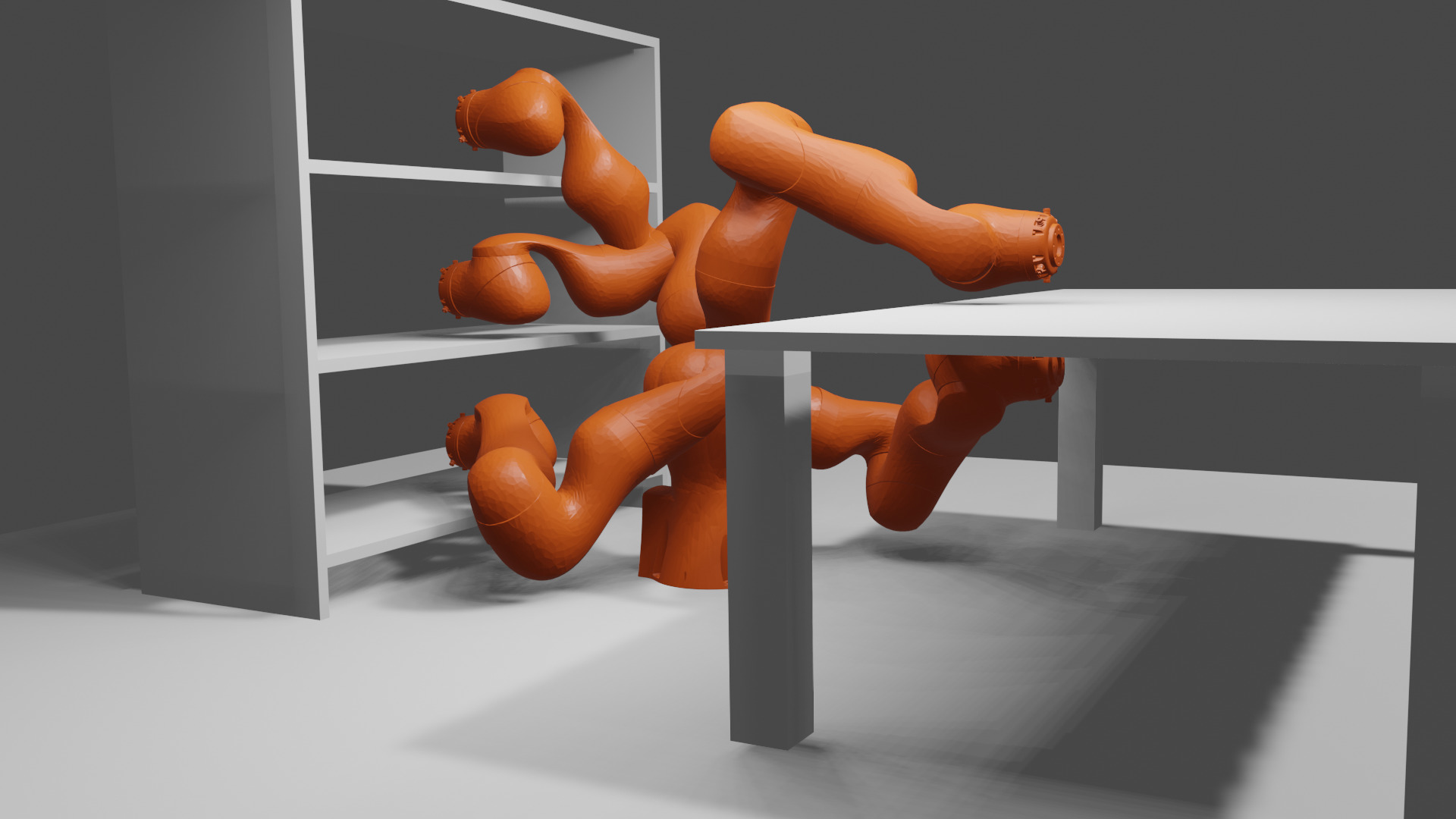}}
    }
    \centerline{
    \subfloat[$\mathcal{P}_{16}$]{\includegraphics[width=0.195\textwidth]{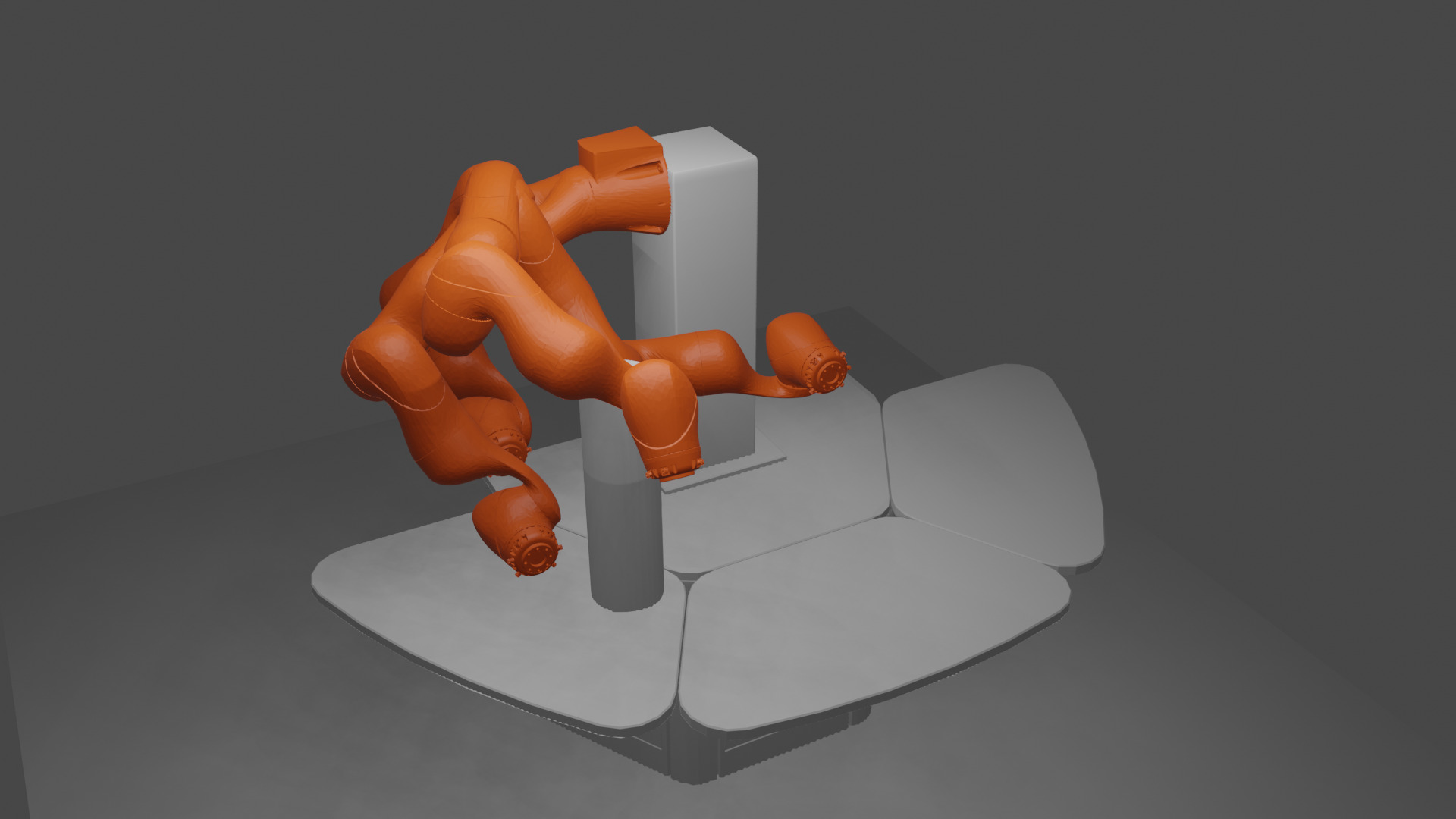}}
    \hfil
    \subfloat[$\mathcal{P}_{17}$]{\includegraphics[width=0.195\textwidth]{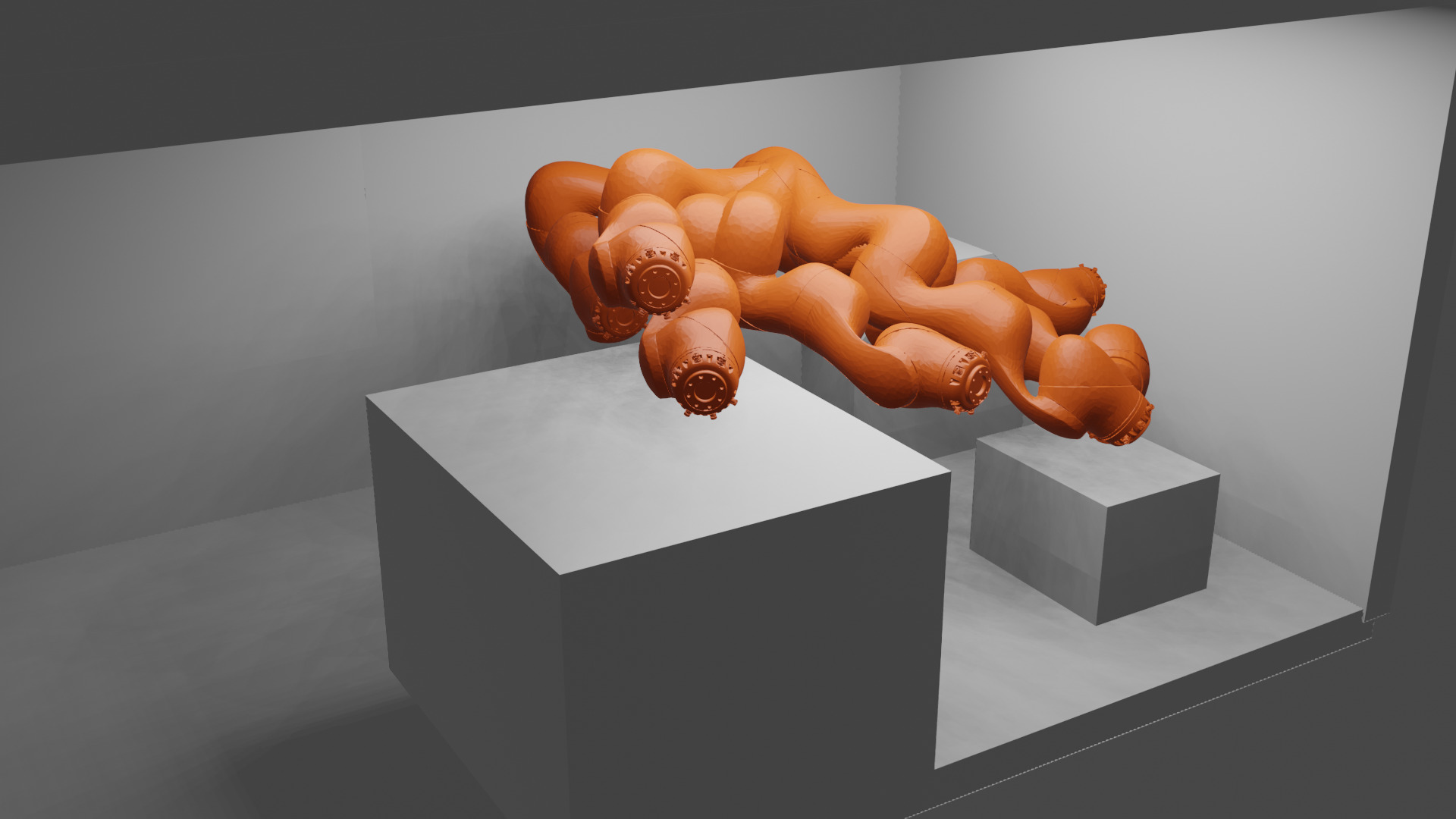}\label{fig:p17}}
    \hfil
    \subfloat[$\mathcal{P}_{18}$]{\includegraphics[width=0.195\textwidth]{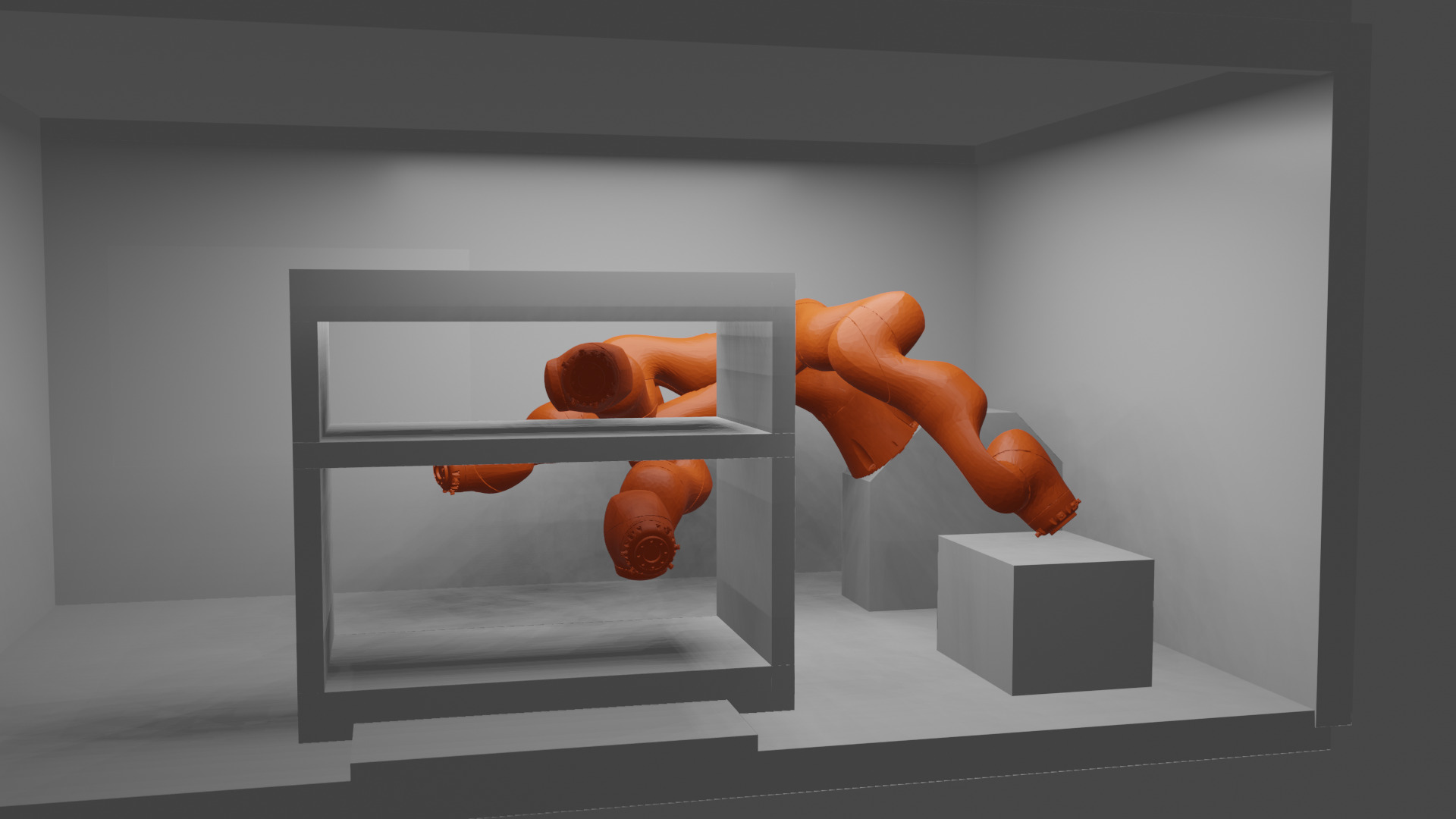}}
    \hfil
    \subfloat[$\mathcal{P}_{19}$]{\includegraphics[width=0.195\textwidth]{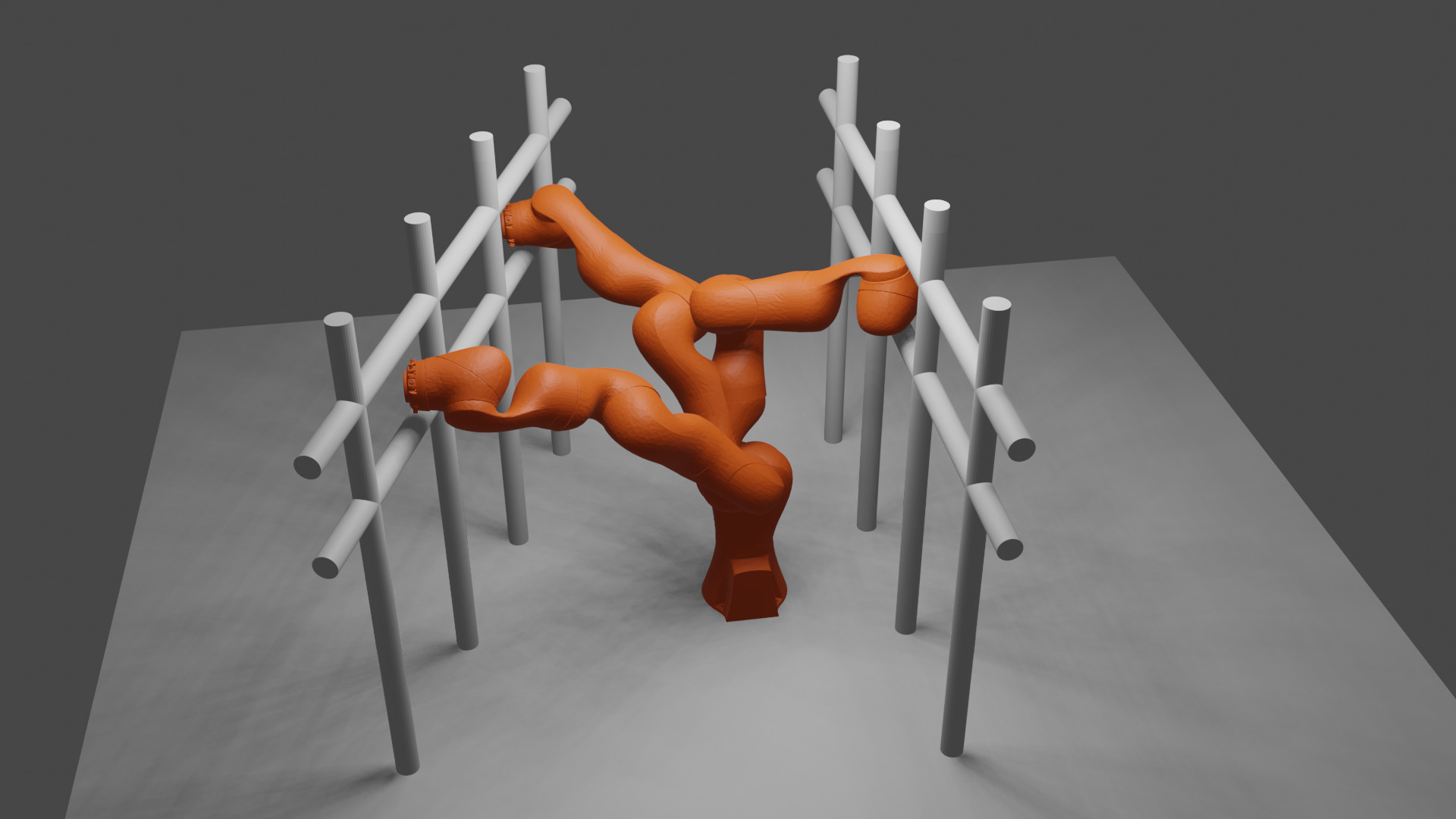}}
    \hfil
    \subfloat[$\mathcal{P}_{20}$]{\includegraphics[width=0.195\textwidth]{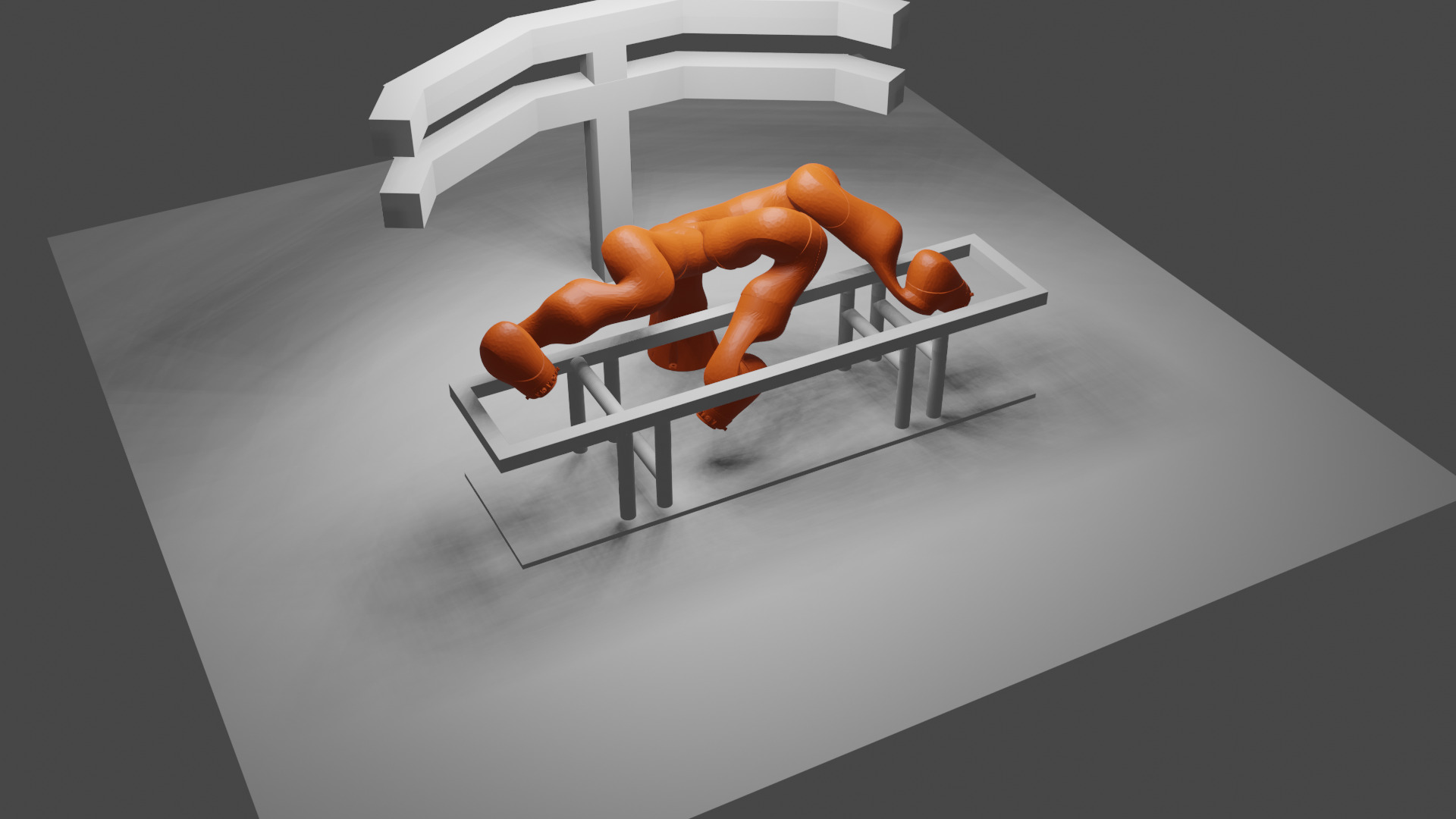}}
    }
    
    \caption{Ten example environments included in the presented benchmark. Some of them contain narrow passages. Others do not include many obstacles and are thus easier to solve.}

    \label{fig:benchmark}
\end{figure*}

Related works utilize a wide range of planning problems and applications for evaluation. However, they are often not made publically available which hinders comparing results to previous works. Moreover, they are usually based on randomized environments or tailored towards a specific problem class. Our contribution is thus to provide a manipulator motion planning benchmark (see Fig. \ref{fig:benchmark}). It consists of 20 environments with several robot configurations per environment. Each environment $e$ can be used to construct a set of possible planning problems $\mathcal{P}_e$. For now, only configurations for the Kuka iiwa manipulator are provided. However, we provide Blender and Coppelia files for each environment which allows for including other manipulators. Furthermore, the benchmark is independent of any planning software. Hence, it can also be used with well-established tools such as the \gls{ompl} benchmark pipeline \cite{MollBenchmarking}. In total, the benchmark contains 214 possible planning problems that can be generated by combining the provided configurations. As it can be seen in Fig. \ref{fig:benchmark}, the provided planning problems contain several narrow passages and thus yield a challenge for most state-of-the-art algorithms. The whole dataset can be downloaded from \href{https://mytuc.org/rybj}{https://mytuc.org/rybj}.

\section{Experiments}
In the following, we conduct two experiments using the previously presented benchmark of planning problems. For all our experiments we use our own RRT*-Connect implementation. In contrast to most implementations in \gls{ompl}, the algorithm is highly parallelized. We restrict number of threads per trial to eight. Throughout our experiments, we optimize the parameters goal bias $b_{goal} \in (0.05,0.75)$ and step size $s \in (\SI{15}{\degree},\SI{135}{\degree})$.
We use the \gls{bohb} implementation provided by Tune \cite{liaw2018tune}. For each problem within a trial, the planning time is restricted to $t_{max} = \SI{20}{\second}$. All experiments are executed on a workstation with two Intel Xeon E5-2670 v3 CPUs, each with 12 processor cores and 64 GB of Ram.

In the following, we first demonstrate the challenge of finding an optimized planner configuration for multiple environments. In a second experiment, we extend the set of considered problems to obtain a distribution of optimal parameters. This allows for extracting a set of suggested parameters depending on the setup.

\subsection{Optimizing on a Single Environment}
\label{sec:exp1}
In the first experiment we investigate weather a planner tuned on one environment can generalize across environments with other characteristics. For this, we used the \gls{bohb} algorithm on the planning problem sets $\mathcal{P}_2$,$\mathcal{P}_3$ and $\mathcal{P}_5$ (see Fig. \ref{fig:benchmark}(a)-(c)). As it can be seen, the three environments differ in the number of obstacles and narrow passages. Moreover, some planning problems within an environment are more challenging than others. In $\mathcal{P}_3$, for example, finding a path to the other side of the arch is more challenging than finding one to a configuration on the same side.
Hence, our first goal is to test the generalization capabilities of an optimized planner. For this, we run \gls{bohb} with 100 trials on all problems of an environment and test the results on all three.
For obtaining a stable loss estimate per environment, we sum up the costs for all problems. The process was repeated five times for each environment $e$ and the 0.7-quantile was used as a loss $L_e$.
We also optimize the parameters using the combined set $\mathcal{P}_{2,3,5}$. In addition, we include the settings suggested by \gls{ompl} for the Kuka iiwa manipulator as a baseline. The results can be seen in Table \ref{tab:exp_1}. The first column describes from which source the optimized parameters $s$ and $b_{goal}$ were taken. Their values can be seen in column three and four. The remaining columns show the mean loss and the 0.7-quantile for all environments. The last column shows the average loss over all environments $\mathcal{P}_2$,$\mathcal{P}_3$ and $\mathcal{P}_5$.

Looking at the optimized parameters, it can be seen that the step size in $\mathcal{P}_2$ has the smallest value which most-likely results from the robot reaching inside the walls. Environment five, on the other hand, is characterized by a very high step size and goal bias because not many obstacles are present within the scene. In consequence, the loss values for the hold-out environments are much higher. Therefore, the optimized parameters are not applicable across all environments. Taking the optimal parameters for $\mathcal{P}_5$, $\mathcal{P}_2$ is solved with more than double the loss compared to using its optimal parameters.
Optimizing for all three environments, as shown in the last column, leads to a trade-off. This inter-class generalization was also reported in \cite{MollHyperplan}. However, this trade-off is highly biased towards $\mathcal{P}_3$. The reason for this is that $\mathcal{P}_3$ is more prone to returning higher planning times and costs. In consequence, the costs and optimized parameters are close to the results for optimizing on $\mathcal{P}_3$. At the same time, it leads to considerably higher costs for $\mathcal{P}_2$. The same can be observed for the \gls{ompl} baseline settings. In the latter case, the parameters work very well on $\mathcal{P}_3$ but deliver poor results on $\mathcal{P}_2$. The combined loss, however is also low.

It can be seen that inter-class optimization of hyperparameters can be challenging due to an optimization bias towards harder problems. When extending the optimization set to even more environments, it is expected that this issue increases. It is thus necessary to identify the modes of the parameter distribution instead of relying on a single setting for all problems. 

\begin{table*}[]
    \caption{results for experiment 1}
    \label{tab:exp_1}
    \renewcommand{\arraystretch}{1.5}
    \centering
    \begin{tabular}{|l||r|r||r|r||r|r||r|r||r|}
    \hline
    \multicolumn{1}{|c||}{} &
    \multicolumn{1}{c|}{$s$} &
    \multicolumn{1}{c||}{$b_{goal}$} &
      \multicolumn{1}{c|}{\begin{tabular}[c]{@{}c@{}}$L_2$ \\ (mean)\end{tabular}} &
      \multicolumn{1}{c||}{\begin{tabular}[c]{@{}c@{}}$L_2$ \\ $(Q_{0.7})$\end{tabular}} &
      \multicolumn{1}{c|}{\begin{tabular}[c]{@{}c@{}}$L_3$ \\ (mean)\end{tabular}} &
      \multicolumn{1}{c||}{\begin{tabular}[c]{@{}c@{}}$L_3$ \\ $(Q_{0.7})$\end{tabular}} &
      \multicolumn{1}{c|}{\begin{tabular}[c]{@{}c@{}}$L_5$ \\ (mean)\end{tabular}} &
      \multicolumn{1}{c||}{\begin{tabular}[c]{@{}c@{}}$L_5$ \\ $(Q_{0.7})$\end{tabular}} &
      \multicolumn{1}{c|}{\begin{tabular}[c]{@{}c@{}}$L_{2,3,5}$ \\ (mean)\end{tabular}} \\ \hline
    $\mathcal{P}_2$ & 0.68 & 0.47     & 89.73  & 99.91  & 371.45 & 428.68 & 17.52 & 18.90 & 159.57\\ \hline
    $\mathcal{P}_3$&2.01&0.18     & 129.25 & 138.72 & 90.76  & 95.82  & 12.79 & 13.19 & 77.60\\\hline
    $\mathcal{P}_5$&2.26&0.59     & 227.66 & 301.23 & 109.20 & 111.36 & 11.68 & 12.31 & 116.16\\\hline
    $\mathcal{P}_{2,3,5}$&1.87&0.08     & 146.93 & 135.78 & 93.43 & 99.88 & 12.28 & 12.73&84.21\\\hline
    OMPL &2.74&0.05& 177.01 & 254.75 & 79.44  & 83.34  & 12.65 & 12.64 &89.70\\\hline
    \end{tabular} 
\end{table*}

\subsection{Cluster Analysis}
\begin{figure} 
  \centerline{
    \includegraphics[]{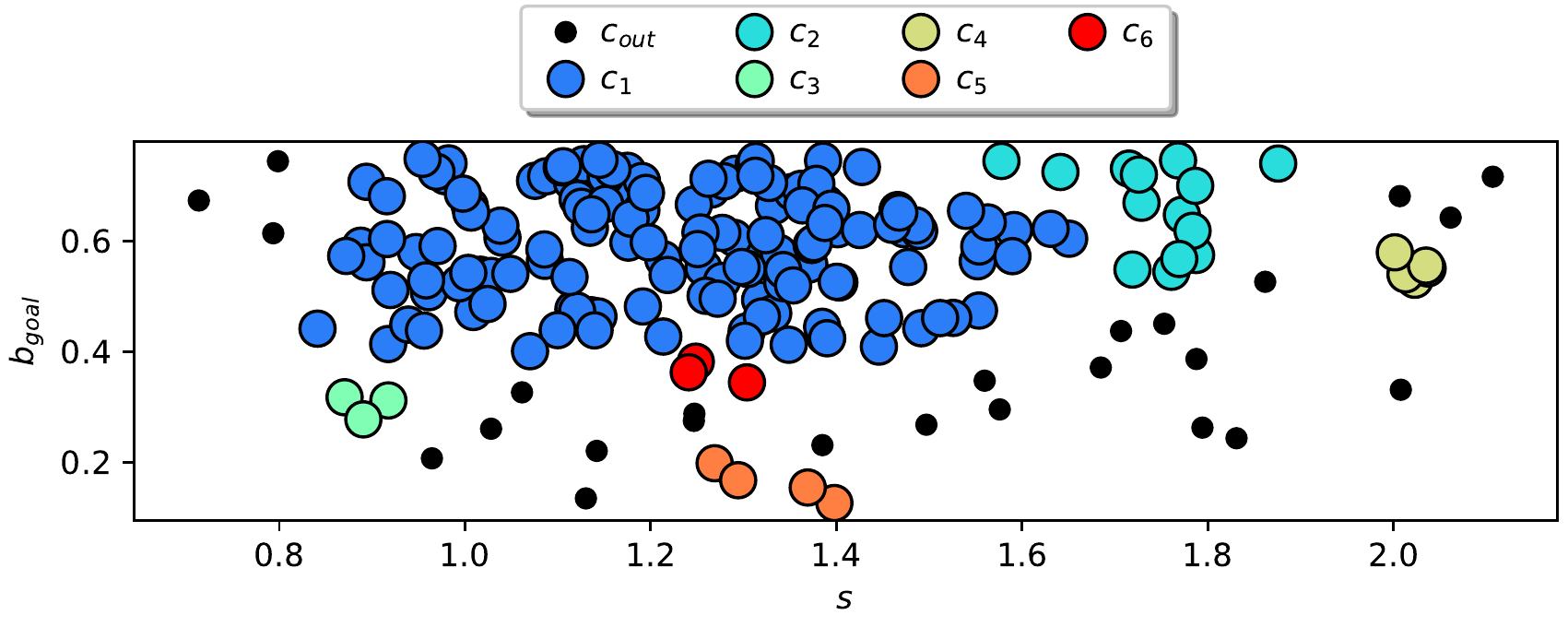}
  }
  
  \caption{Results of the clustering analysis defining a set of six parameter settings. Samples labeled as outliers are displayed as black dots.}
  \label{fig:exp2_cluster}
\end{figure}

In the next step, we provide insights in the distribution of optimal parameters for a large set of planning problems. The goal is to define regions of optimal parameters and connect them to the characteristics of the planning problem. For this, we use the three environments from \ref{sec:exp1} as a test set. The remaining 17 environments and their 196 planning problems define the data basis for our analysis. We run \gls{bohb} with 100 trials for each of the problems and record the best step size and goal bias. For a more stable estimate, we solve each problem five times and use the loss function from \eqref{eq:costs_final}. In a subsequent step, we run a cluster analysis on the resulting 196 data points. For this, we use the scikit-learn \cite{scikit-learn} implementation of the \gls{dbscan} algorithm \cite{ester199dbscan}.
Using the metric defined in \eqref{eq:metric}, we set the algorithm's parameters to $\varepsilon  = 0.15$ and $\textit{min\_samples}=3$. These settings were chosen with the goal of not obtaining too many outliers or only a single cluster. The resulting clusters can be seen in Fig. \ref{fig:exp2_cluster}. The final number of outliers is 25 and the number of samples per cluster is given in Table \ref{tab:exp_2}. In addition, we provide a list of planning problems per cluster in the attachment.

\begin{figure} 
  \centerline{
    \includegraphics[]{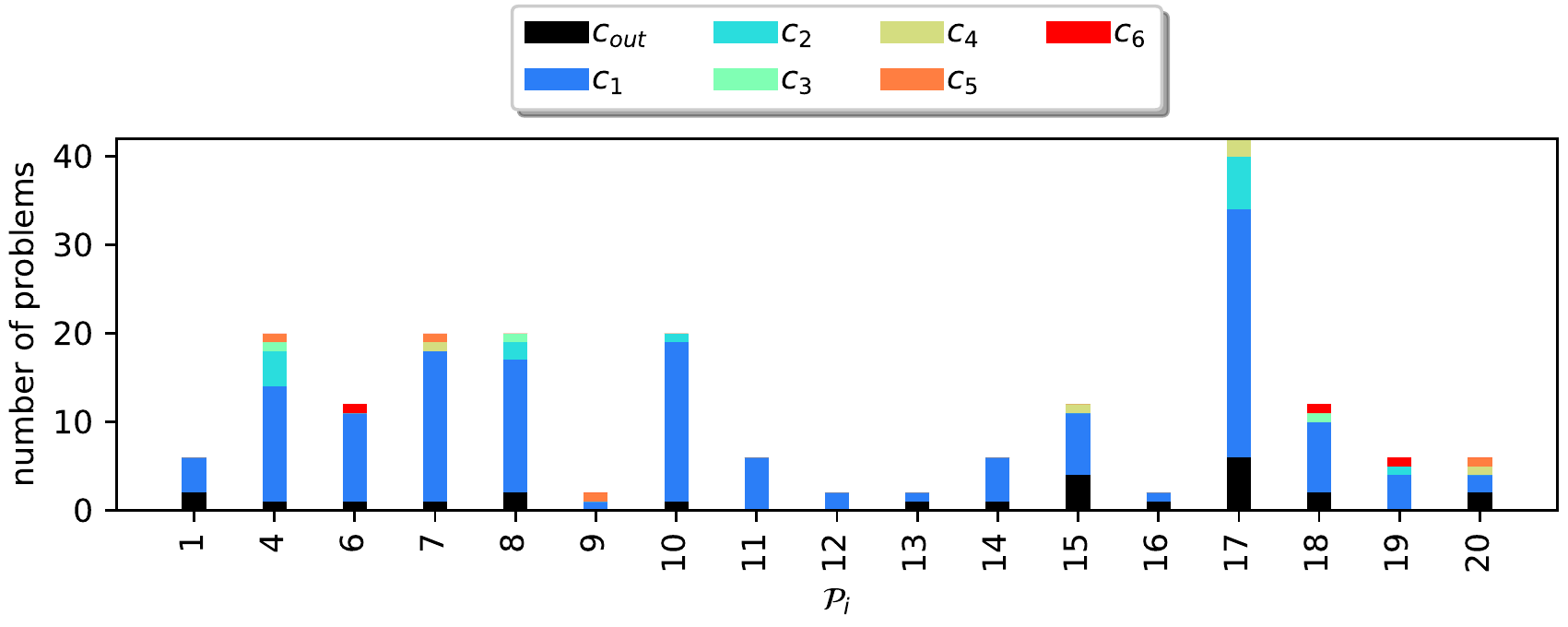}
    }

  \caption[]{Number of clustered problems per class. It can be seen that the optimized parameters vary even within the same problem class. However, settings from $c_1$ are applicable for most of the problems.}
  \label{fig:exp2_bar}
\end{figure}

\begin{figure}[ht]
  \centerline{
    \subfloat[Example for $c_1$]{\includegraphics[width=0.48\columnwidth]{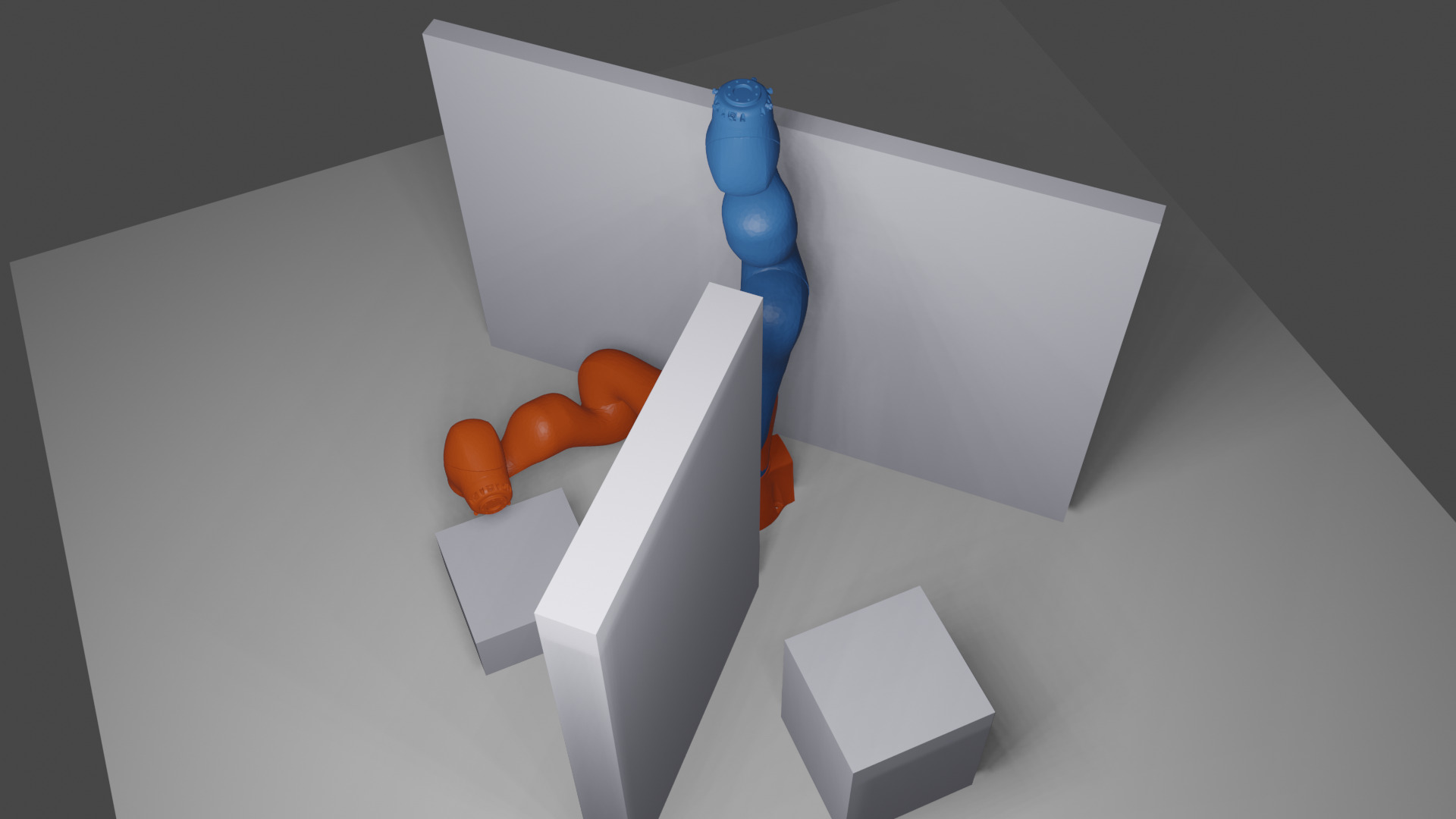}\label{fig:clusters_examples_a}} 
    \hfil
    \subfloat[Example for $c_2$]{\includegraphics[width=0.48\columnwidth]{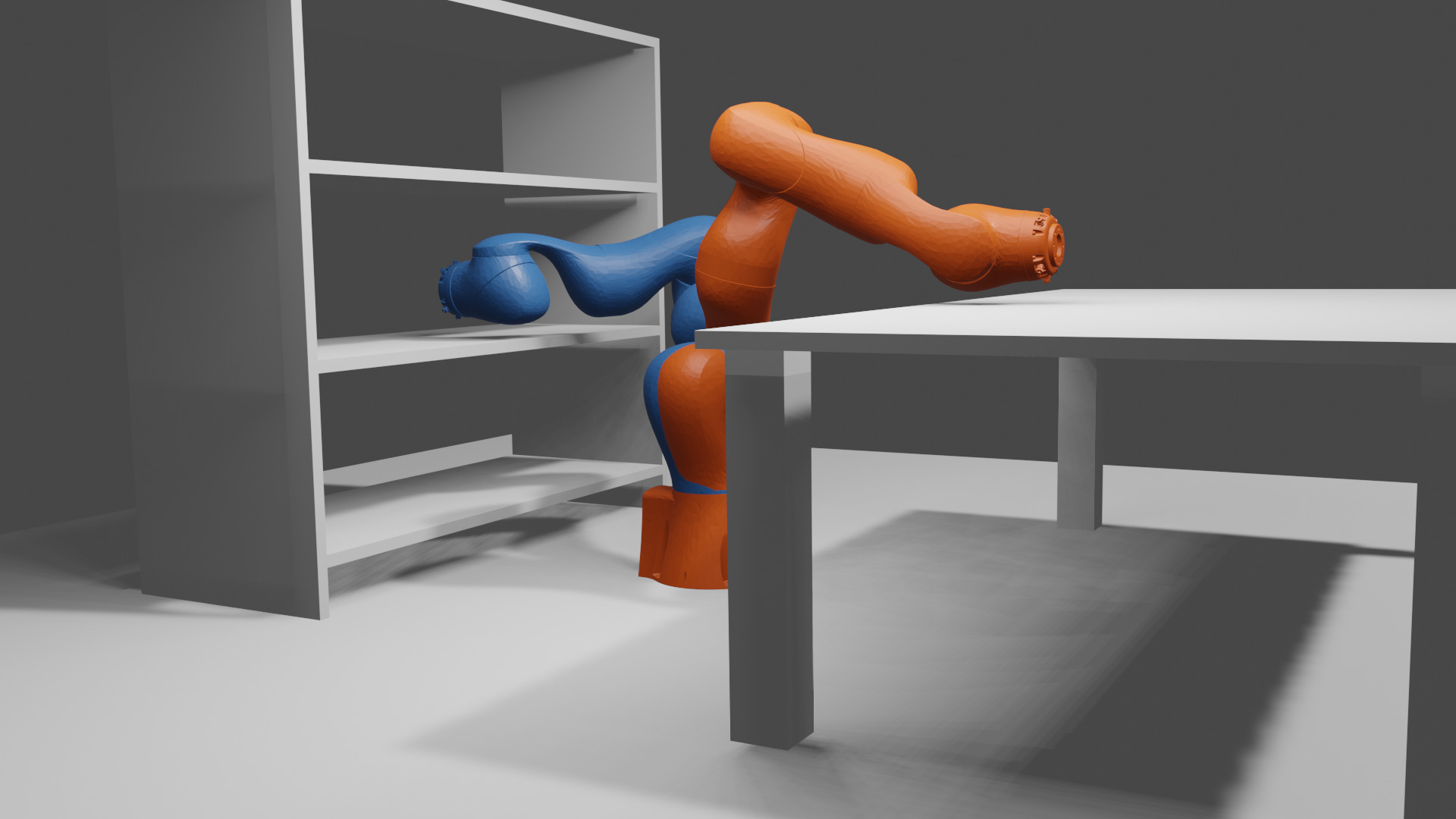}\label{fig:clusters_examples_b}}
  }
  \centerline{
    \subfloat[Example for $c_3$]{\includegraphics[width=0.48\columnwidth]{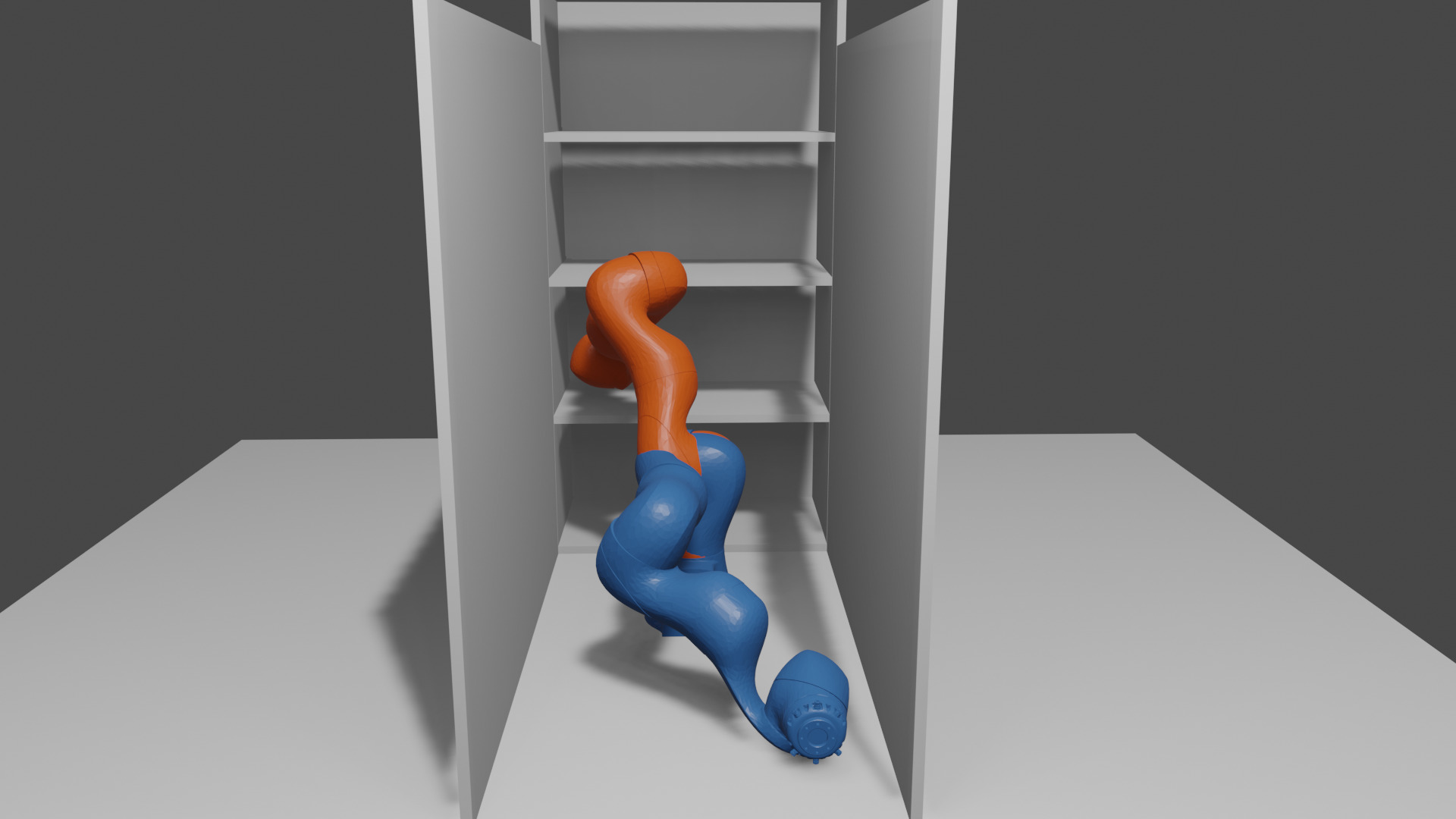}\label{fig:clusters_examples_c}}
    \hfil
    \subfloat[Example for $c_4$]{\includegraphics[width=0.48\columnwidth]{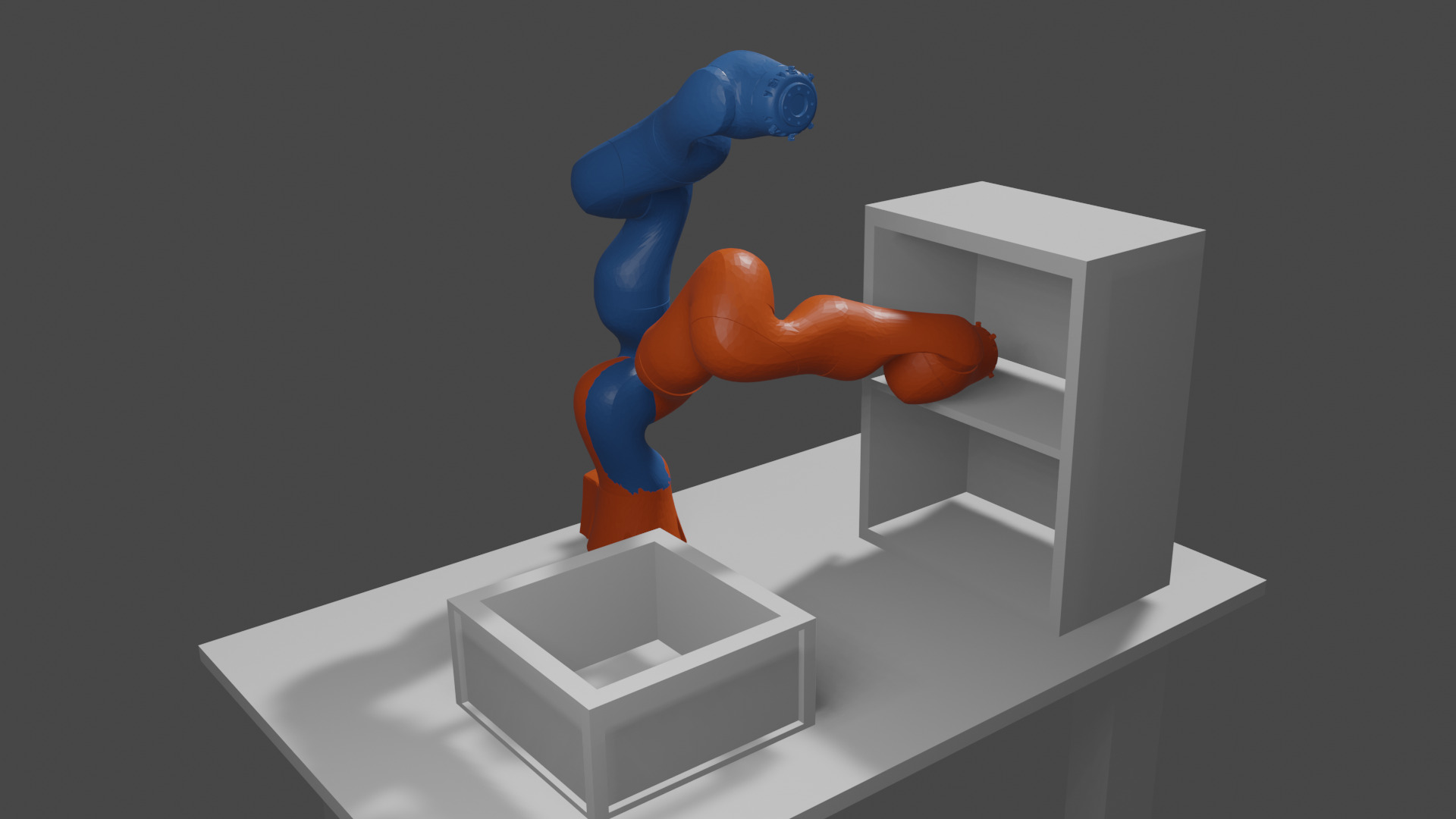}\label{fig:clusters_examples_d}}
  }

  \centerline{
    \subfloat[Example for $c_5$]{\includegraphics[width=0.48\columnwidth]{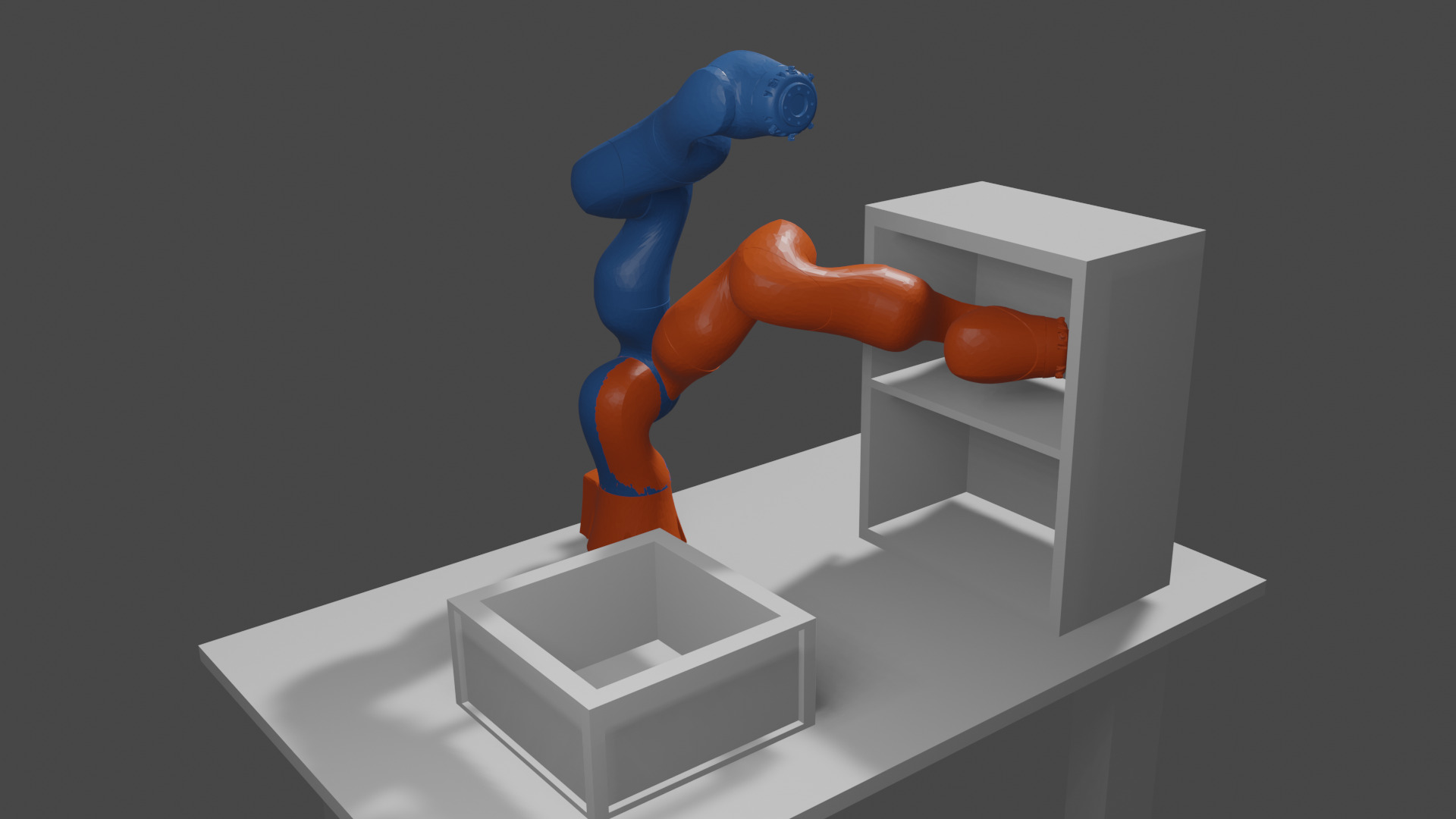}\label{fig:clusters_examples_e}}
    \hfil
    \subfloat[Example for $c_6$]{\includegraphics[width=0.48\columnwidth]{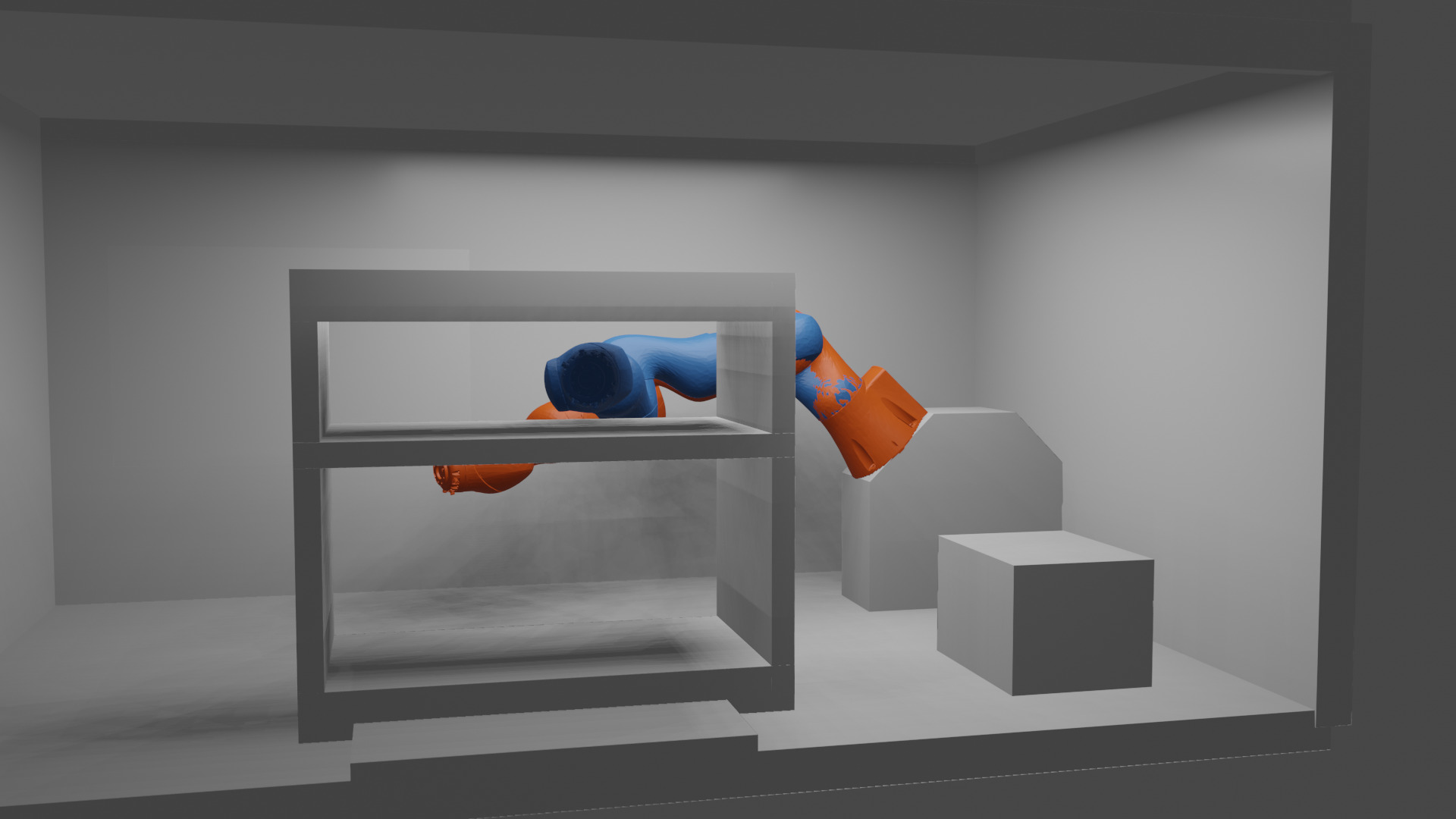}\label{fig:clusters_examples_f}}

  }  
  \caption{Example planning problems per cluster. Start configurations are shown in orange and goals in blue.}
  \label{fig:clusters_examples}
\end{figure}

It can be seen that the obtained parameters are relatively widely spread across most of the parameter range. This stems from the fact that the data is based on a stochastic motion planning procedure. We also use a limited set of trials during optimization.  It is thus necessary to consider the resulting data points as noisy and use appropriate clustering algorithms such as \gls{dbscan}.
Furthermore, a distribution of the individual planning problems per cluster and environment is given in Fig. \ref{fig:exp2_bar}. It can be seen that settings from $c_1$ are most dominant in all environments. This results from the fact that most environments also contain planning problems which are less complex. In $\mathcal{P}_{17}$, for instance, a significant amount of problems involves moving the manipulator over the small box and away from the narrow passage (see Fig. \ref{fig:p17}).
Considering the individual problems, however, one can see some similarities. A limited set of examples is provided in Fig. \ref{fig:clusters_examples} and a complete list can be found with the benchmark files. The cluster $c_1$ contains most data points and its center thus represents a good general setting for medium complexity. This means it is well-suited for situations where the robot does not have to reach far inside a narrow passage. In contrast, cluster $c_2$ is characterized by a larger step size and suited for less cluttered environments. When operating in mostly free spaces and not reaching into narrow passages, $c_4$ is recommended. In contrast, the problems in cluster $c_5$ are characterized by narrow passages and hard-to-reach goal configurations. Comparing Fig. \ref{fig:clusters_examples_d} and Fig. \ref{fig:clusters_examples_e}, one can see that reaching a bit further inside the shelf already requires adapting the parameters from $c_4$ to $c_5$. The remaining setting in $c_3$ and $c_6$ can be seen as an extension of $c_1$ since they are close and not supported by many data points. We thus recommend considering them for problems of medium complexity as well.

Finally, we use the test problems $\mathcal{P}_2, \mathcal{P}_3$ and $\mathcal{P}_5$ to validate the feasibility of the obtained parameters. The values for the cluster centers as well as the number of samples per cluster are shown in Table \ref{tab:exp_2}. In the second part of the table the results of using these parameters are shown. For this, we used the same test setup as in \ref{sec:exp1}. It can be seen that the challenging set $\mathcal{P}_2$ can be solved most efficiently with the parameter settings from $c_5$ whereas $c_2$ performs poorly. In contrast, $c_2$ is well-suited for  $\mathcal{P}_5$ since the little amount of obstacles allow for larger step sizes and goal biases. In environments with trivial as well as harder problems, such as $\mathcal{P}_3$, a reduced goal bias and increased step size such as in $c_4$ works best on average. On the other hand, using the center of $c_3$ is not recommended because small values for $b_{goal}$ and $s$ lead to longer planning times when solving the easier problems at one side of the arch.

\begin{table}[]
  \caption{Cluster center and their performance on the test set}
  \label{tab:exp_2}
  \renewcommand{\arraystretch}{1.5}
  \centering
  \begin{tabular}{|l|r|r|r||r|r|r|}
  \hline
  \multicolumn{1}{|c|}{cluster} &
  \multicolumn{1}{c|}{$s^*$} &
  \multicolumn{1}{c|}{$b_{goal}^{*}$} &
  \multicolumn{1}{c||}{$|c_i|$} &
  \multicolumn{1}{c|}{\begin{tabular}[c]{@{}c@{}}$L_2$ \\ $(Q_{0.7})$\end{tabular}} &
  \multicolumn{1}{c|}{\begin{tabular}[c]{@{}c@{}}$L_3$ \\ $(Q_{0.7})$\end{tabular}} &
  \multicolumn{1}{c|}{\begin{tabular}[c]{@{}c@{}}$L_5$ \\ $(Q_{0.7})$\end{tabular}} \\ \hline
  $c_1$ & 1.23  & 0.59  & 142 & 263.25& 129.37& 16.90\\ \hline
  $c_2$ & 1.74  & 0.66  & 14 & 469.36& 109.47& \textbf{11.88}\\ \hline
  $c_3$ & 0.89  & 0.30  & 3 & 131.47& 192.46& 19.31\\ \hline
  $c_4$ & 2.02  & 0.55  & 5 & 152.27& \textbf{93.56}& 12.00\\ \hline
  $c_5$ & 1.33  & 0.16  & 4 & \textbf{82.16}& 113.34& 14.40\\ \hline
  $c_6$ & 1.26  & 0.36  & 3 & 116.20& 120.05& 17.89\\ \hline
  \end{tabular} 
\end{table}

\addtolength{\textheight}{-0.6cm}

\section{Discussion}
In this work, we have presented a novel benchmark for manipulator motion planning. The dataset contains a set of 20 planning environments and 214 planning problems in total. The main contribution of this work, however, is optimizing the parameters of the RRT*-Connect algorithm. First, we investigated weather a planner optimized on one environment can be employed in other setups. In contrast to previous works, we chose environments with inherently different characteristics and complexity. Such settings are closer to dynamic environments such as shared workspaces. We show that this is not always possible. Moreover, the planner's performance can even drop when optimized on a different setup. In some cases even below the baseline.
When jointly optimizing the planner on all three environments, we achieved slightly better results than with the baseline setting. However, they do not outperform the individually optimized planners. Moreover, the optimization was tailored towards $\mathcal{P}_3$ which is most prone to deliver high loss values. The optimization procedure thus leads to a reasonable performance in general but gives up on the advantages in some setups. Depending on the chosen optimization set, such a general parameter setting could even lead to very poor performance in certain cases which should be avoided in \gls{hri} applications.

For this reason, we analyzed the distribution of optimal parameter settings for 196 problems across 17 environments. The resulting data, however, does not show a very clear cluster structure due to the stochastic planning process and the limited number of optimization trials. Due to these noisy data points, we used the established \gls{dbscan} algorithm together with an adapted distance metric to identify six clusters. This number, however, is influenced by the parameter settings of the algorithm and can vary. When choosing a setting for a concrete use-case, we recommend $c_1, c_3$ and $c_6$ for environments with medium clutter. The robot should also not have to reach far inside narrow passages. For less-densely cluttered workspaces, $c_2$ can be considered. The settings for $c_5$ are recommended for complex problems with narrow passages and $c_4$ for trivial problems.
We utilized the hold-out test set from our first analysis to evaluate the suggested parameters. The results are equally good to optimizing the planner directly on the test environments. The presented set of parameters can thus help practitioners setting up their planner given the category of planning problems.

However, our approach is limited in some ways. First, we use a highly parallelized custom version of the RRT*-Connect algorithm. Hence, our method explores the search space much more rapidly than single-threaded planners. It is thus not clear yet how well the suggested settings transfer to other implementations or even algorithms. In addition, it is possible to cluster the obtained data differently. This would lead to a different number of outliers and clusters. Moreover, this work only provides informed guidelines for setting parameters optimally. It is not possible yet to directly extract them from the concrete planning problem. For future work, we are interested in extracting relevant characteristics from the environment by learning a latent representation and assigning parameters automatically.




\FloatBarrier 
\section*{ACKNOWLEDGMENT}
Funded by the Deutsche Forschungsgemeinschaft (DFG, German Research Foundation) – Project-ID 416228727 – SFB 1410


\bibliographystyle{IEEEtran}
\bibliography{IEEEabrv,literature}

\end{document}